%% file: main.tex
\PassOptionsToPackage{table}{xcolor}
\documentclass{article}
\usepackage{iclr2027_conference,times}
\usepackage[T1]{fontenc}
\usepackage[hyphens]{url}
\usepackage{graphicx}
\usepackage{caption}
\usepackage{booktabs}
\usepackage{multirow}
\usepackage{array}
\usepackage{placeins}
\usepackage{amsfonts}
\usepackage{amsmath}
\usepackage[hidelinks]{hyperref}
\providecommand{\checkmark}{\ensuremath{\surd}}
\iclrfinalcopy

\title{\textsc{TSAE}: Structured Sparse Autoencoders for Interpreting Time-Series Forecasting Models}
\author{Baoxi Liu \and Yi Xie}

\begin{document}

\maketitle

\begin{abstract}
Time-series forecasting informs critical decisions in energy dispatch, industrial
operations, and environmental monitoring; understanding the patterns models rely on is
essential for assessing reliability and identifying failures. Input attribution
identifies important variables and time segments but offers limited insight into
internal features, while standard sparse autoencoder (SAE) objectives do not
directly constrain cross-variable structure or temporal continuity. We introduce
\textsc{TSAE}, a structured sparse autoencoder for forecasting representations that
decomposes hidden states into individually inspectable features. \textsc{TSAE} organizes
cross-variable structure through shared and variable-routed private dictionaries,
separates feature detection from magnitude estimation with gated encoding, and
constrains neighboring sparse-code changes according to raw-segment similarity. These
mechanisms support analysis of variable context, activation strength, and temporal
evolution. Forecast-consistency fine-tuning further improves preservation of the frozen
forecaster's outputs. The accompanying \textsc{TSEval} protocol separately audits
fidelity, feature coherence, and physical calibration to ground feature interpretation.
In three-seed experiments with frozen PatchTST on ETTh1, ETTh2, and ETTm1, \textsc{TSAE}
achieves the lowest hidden-state reconstruction error, normalized
forecast-reconstruction error (NFRE), and feature transition rate among five SAEs at
comparable per-token activity. NFRE decreases by 4.3--26.4\% relative to the next-best
mean. Dataset-dependent tradeoffs in selectivity, physical correlation, and calibration
show that fidelity and temporal-stability gains require independent semantic validation
to support feature interpretation.
\end{abstract}

\section{Introduction}

Time-series interpretability is essential for deploying forecasting models in high-stakes domains such as energy, climate, healthcare, and finance. In these settings, prediction errors can lead to substantial real-world costs, making model reliability crucial.

Most widely used time-series interpretability methods explain forecasts at the input level. Saliency maps, occlusion analysis, Shapley values, dynamic masking, and counterfactual explanations attribute predictions to variables, timesteps, or perturbed subsequences~\citep{sundararajan2017axiomatic,bento2021timeshap,crabbe2021dynamask}. However, input-level attribution alone does not probe the latent features learned by the forecasting backbone, making it difficult to assess whether the model encodes semantically coherent, physically grounded, and forecast-relevant temporal patterns.

Recently, Sparse Autoencoders (SAEs) have become an increasingly used approach for interpreting hidden representations in large-scale models. By decomposing hidden states into sparse linear combinations of learned feature directions~\citep{bricken2023monosemanticity,gao2024scaling}, SAEs can help identify recurring internal concepts beyond input-level attribution. While established practices in language model interpretability---such as assessing reconstruction fidelity, measuring sparsity, and analyzing top-activating examples---are well-defined, their application to time-series remains underexplored. This gap highlights a significant opportunity to leverage SAEs for time-series interpretability, potentially exposing reusable, semantically meaningful hidden features within forecasting backbones.

\begin{figure}[t]
\centering
\includegraphics[width=\linewidth]{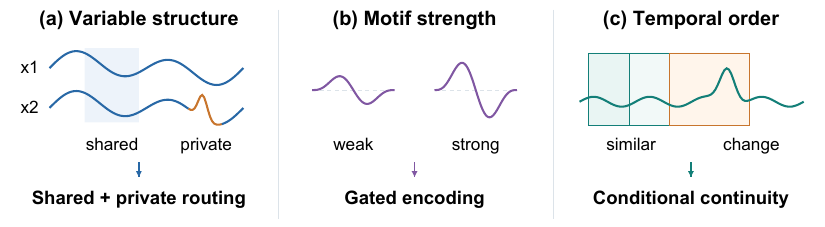}
\caption{Three time-series properties motivate \textsc{TSAE}: (a) shared motifs and variable-specific structure motivate shared/private routing; (b) motif strength motivates separating detection from magnitude; (c) similar adjacent patches of one variable motivate code continuity, with weaker coupling across real changes. Traces are schematic.}
\label{fig:intro-motivation}
\end{figure}

However, standard SAEs are not designed to explicitly model the time-series structure illustrated in Figure~\ref{fig:intro-motivation}. In multivariate forecasting, dependencies across variates can make shared temporal patterns relevant~\citep{liu2024itransformer}, while patch-based models construct neighboring tokens from overlapping local segments~\citep{nie2023patchtst}. When a backbone exposes variable-indexed, ordered tokens, a useful time-series SAE should distinguish meaningful cross-variable sharing from uncontrolled mixing and encourage stable feature identities across adjacent patches when their raw segments are similar. Separately, because sparsity penalties can shrink activation magnitudes~\citep{rajamanoharan2024improving}, it is important to audit rather than assume whether sparse activations remain calibrated to local signal variation. A flat reconstruction--sparsity objective does not directly encourage these properties.

Beyond learning sparse features, time-series interpretability also requires evaluating whether those features are meaningful for forecasting. Reconstruction loss and top-activating examples indicate that a dictionary can preserve hidden states and expose inspectable patterns, but they do not show whether a feature is coherent across variables, calibrated to signal motifs, or relevant to forecasting. We therefore need an evaluation protocol that separates reconstruction fidelity, feature purity, physical calibration, forecast sensitivity, and input-level baselines.

To address these limitations, we propose \textsc{TSAE} (Time-Series Structured Sparse Autoencoder), a structured sparse autoencoder for interpreting time-series forecasting representations. \textsc{TSAE} combines shared and variable-routed private dictionaries with mechanisms for magnitude preservation and temporal stability, encouraging a separation between shared temporal motifs and variable-conditioned residual structure. Complementing the model, we introduce \textsc{TSEval}, a capability-graded multi-axis diagnostic protocol that evaluates sparse features beyond reconstruction, including purity, physical calibration, and forecast intervention sensitivity.

Our contributions are:
\begin{itemize}
\item We introduce \textsc{TSAE}, a structured SAE for variable-indexed and time-ordered forecasting representations, together with an inductive-bias analysis explaining how shared/private dictionaries, magnitude preservation, and temporal regularization are designed to address limitations of flat reconstruction--sparsity objectives.
\item We develop \textsc{TSEval}, a six-axis evaluation protocol and adapter toolkit for auditing sparse features in terms of reconstruction, purity, physical calibration, forecast sensitivity, hidden-space intervention, and input-level baselines.
\item We validate \textsc{TSAE} with active-matched comparisons across multiple datasets, forecasting backbones, and independent random seeds, together with component ablations that examine routing, gating, and temporal continuity.
\end{itemize}

Related work is discussed in Appendix~\ref{sec:related-work}.

\section{\textsc{TSAE}: Structured Sparse Autoencoder}
\label{sec:tsae-method}


\begin{figure}[htbp]
\centering
\includegraphics[width=0.98\textwidth]{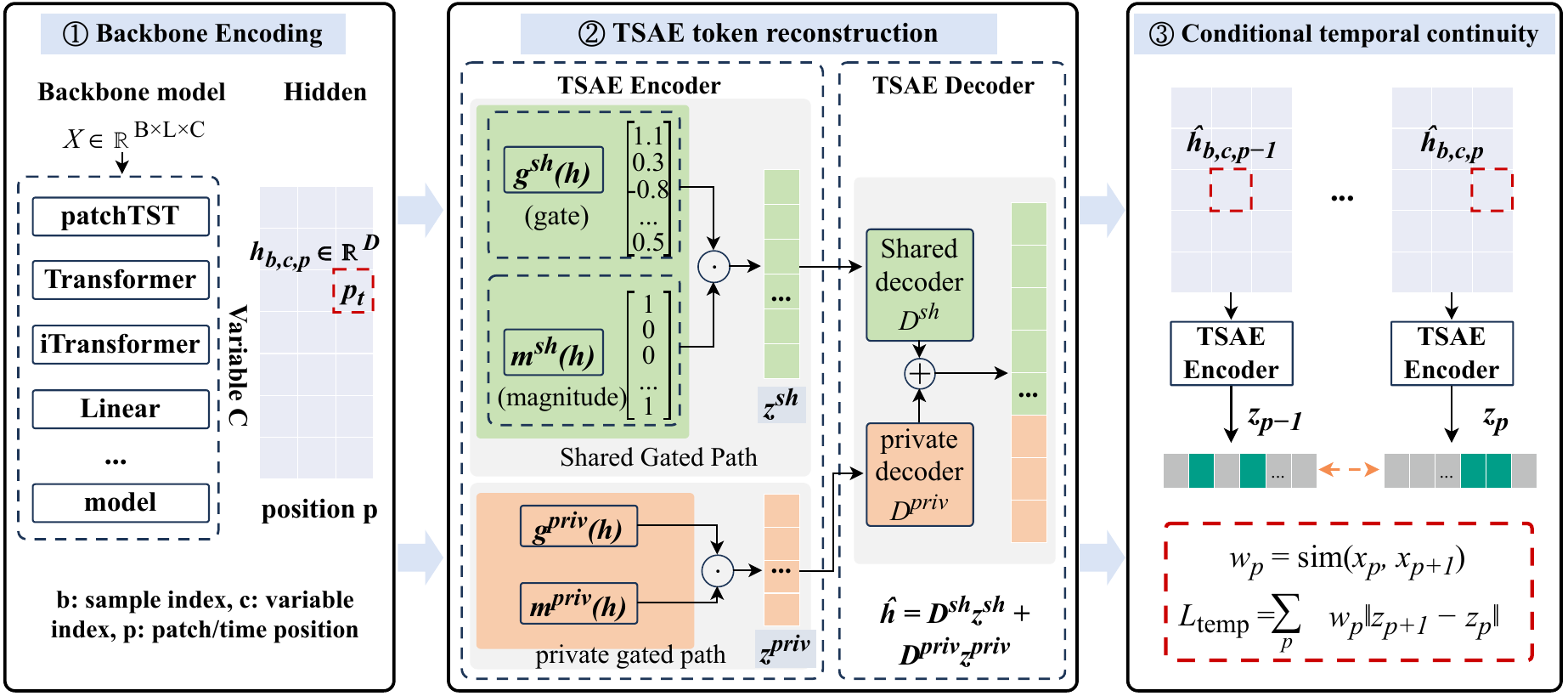}
\caption{\textsc{TSAE} model framework. Raw time-series inputs are first encoded by a forecasting backbone into hidden tokens. For each token, the \textsc{TSAE} encoder produces shared and variable-routed private gated sparse codes, and the decoder reconstructs the hidden token from the corresponding shared and private dictionaries. Conditional temporal continuity is applied to adjacent sparse codes, with the regularization strength determined by the similarity of their raw temporal segments.}
\label{fig:model-framework}
\end{figure}

\textsc{TSAE} is a structured sparse autoencoder inserted at the representation level of a forecasting model. A raw multivariate sequence is first encoded by a forecasting backbone into hidden tokens \(h_{b,c,p}\in\mathbb{R}^{D}\), and \textsc{TSAE} then learns an overcomplete sparse code and reconstructs the hidden token, not the raw input series. We do not assume that hidden states admit a unique decomposition into ground-truth factors; instead, we use three task-specific inductive biases: variable structure, magnitude preservation, and temporal structure. These biases motivate shared/private variable routing, gated detection--magnitude separation, and conditional temporal continuity, respectively.

\paragraph{Shared/private variable routing.}
Let \(h_{b,c,p}\in\mathbb{R}^D\) denote a hidden token from sample \(b\), variable \(c\), and token position \(p\). A flat SAE maps \(h_{b,c,p}\) to an overcomplete sparse vector \(z\in\mathbb{R}^{K}\) and decodes it back to \(\hat h_{b,c,p}\in\mathbb{R}^{D}\) with one shared dictionary. Cross-variable activation is not inherently problematic when a feature consistently captures the same temporal motif across variables; the difficulty is that the same dictionary can also reuse one sparse direction for different motifs under different variable contexts.

\textsc{TSAE} therefore reconstructs each token with a shared component and a variable-routed private component:
\begin{equation}
\label{eq:shared-private-reconstruction}
\hat h =
D^{\mathrm{sh}} z^{\mathrm{sh}}
+ D^{\mathrm{priv}}_{c} z^{\mathrm{priv}}
+ b,
\end{equation}
where \(z^{\mathrm{sh}}\in\mathbb{R}^{K_{\mathrm{sh}}}\), \(z^{\mathrm{priv}}\in\mathbb{R}^{K_{\mathrm{priv}}}\), \(D^{\mathrm{sh}}\) is shared across variables, and \(D^{\mathrm{priv}}_{c}\) is selected by the variable index \(c\). Thus the SAE reconstruction step becomes \(\mathbb{R}^{D}\rightarrow\mathbb{R}^{K_{\mathrm{sh}}+K_{\mathrm{priv}}}\rightarrow\mathbb{R}^{D}\) for each routed token, with private capacity kept compact. The split is learned rather than assigned by a hard rule: each token can activate both paths, while reconstruction, sparsity, and limited private capacity encourage recurring cross-variable motifs to use shared features and variable-conditioned residual structure to use private features. This induces, rather than guarantees, a separation between meaningful sharing and uncontrolled mixing.

\paragraph{Gated detection and magnitude.}
The gated step occurs inside the encoder when \(h\in\mathbb{R}^{D}\) is lifted into the overcomplete sparse space. Each shared or private feature first computes a learned direction and then applies separate gate and magnitude transformations to that direction:
\begin{equation}
\label{eq:gated-code}
z = \mathbf{1}[g(h)>0]\odot m(h),
\end{equation}
Here \(g(h)\) detects whether the feature is present, \(m(h)\) is a nonnegative magnitude estimate, and \(\odot\) denotes elementwise multiplication.
If the gate is closed, the feature activation is zero; if it is open, the activation takes the magnitude value. \textsc{TSAE} applies this parameterization to both the shared path, producing \(z^{\mathrm{sh}}\), and the variable-routed private path, producing \(z^{\mathrm{priv}}\). The sparsity penalty is applied to gate activations rather than directly to magnitudes. This follows the Gated SAE idea of separating feature selection from feature magnitude, but adapts it to time-series interpretation: the model is encouraged to open fewer features, while the magnitude path can still preserve feature strength for comparison with physical motifs such as slope or volatility. Excessive sparsity-driven shrinkage of magnitudes would weaken the link between sparse activations and signal strength.

\paragraph{Conditional temporal continuity.}
The temporal term is applied after the encoder has produced sparse codes, not as a separate reconstruction path. For two neighboring ordered temporal tokens from the same sample and variable, \textsc{TSAE} first computes \(z_{b,c,p}=[z^{\mathrm{sh}}_{b,c,p},z^{\mathrm{priv}}_{b,c,p}]\) and \(z_{b,c,p+1}\) through the same shared/private gated encoder. If the corresponding raw tokens are similar, abrupt changes in sparse feature identity can reflect feature flicker rather than meaningful temporal change; if the raw signal contains spikes or regime changes, sparse features should be allowed to change. \textsc{TSAE} therefore applies temporal smoothing conditionally: feature differences are penalized strongly only when the corresponding raw tokens are similar.

For adapters that expose multiple ordered temporal tokens per variable, we build a chain graph from adapter metadata rather than inferring adjacency from hidden-space similarity. Tokens are grouped by sample and variable, sorted by token position, and adjacent positions are connected. We then apply a weighted graph total variation:
\begin{equation}
\label{eq:temporal-continuity}
L_{\mathrm{temp}}
= \mathbb{E}_{b,c,p}\left[
w_{b,c,p}
\left\|z_{b,c,p+1}-z_{b,c,p}\right\|_1
\right].
\end{equation}
The weight \(w_{b,c,p}\in[0,1]\) is an RBF similarity between adjacent locally normalized raw tokens:
\begin{equation}
\label{eq:raw-patch-similarity}
w_{b,c,p}
=
\exp\left(
-\frac{\mathrm{dist}(\bar x_{b,c,p+1},\bar x_{b,c,p})^2}
{2\sigma^2}
\right),
\end{equation}
Here \(\bar x\) denotes raw patches normalized jointly within each sample--variable chain, \(\mathrm{dist}(u,v)=\|u-v\|_2/\sqrt{\ell}\) is the normalized Euclidean distance between two segments of length \(\ell\), and the PatchTST headline runs use \(\sigma=1\).
Thus, similar raw tokens yield large weights and strongly discourage unnecessary feature switching, whereas dissimilar tokens yield small weights and allow activations to change for spikes or regime transitions. For adapters without ordered temporal edges, such as single-token or non-temporal-token representations, the temporal graph is empty and \(L_{\mathrm{temp}}=0\).

\paragraph{Overall objective.}
The first training stage jointly optimizes the three mechanisms with a compact objective:
\begin{equation}
\label{eq:base-objective}
\begin{aligned}
L_{\mathrm{base}}={}&L_{\mathrm{recon}}+L_{\mathrm{sp}}
+\lambda_{\mathrm{gate}}L_{\mathrm{gate}}\\
&+\lambda_{\mathrm{temp}}L_{\mathrm{temp}}+L_{\mathrm{aux}}.
\end{aligned}
\end{equation}
Here \(L_{\mathrm{recon}}\) reconstructs forecasting hidden tokens and \(L_{\mathrm{sp}}=\lambda_{\mathrm{shared}}\,\overline a^{\mathrm{sh}}+\lambda_{\mathrm{priv}}\,\overline a^{\mathrm{priv}}\), where \(\overline a^{\mathrm{sh}}\) and \(\overline a^{\mathrm{priv}}\) are the mean nonnegative gate signals in the shared and private paths. \(L_{\mathrm{gate}}\) is the gate-reconstruction auxiliary loss, and \(L_{\mathrm{temp}}\) is the conditional temporal continuity term. The auxiliary term is \(L_{\mathrm{aux}}=\lambda_{\mathrm{balance}}L_{\mathrm{balance}}+\lambda_{\mathrm{orth}}L_{\mathrm{orth}}\): \(L_{\mathrm{balance}}\) discourages private-feature usage from collapsing within a variable, while \(L_{\mathrm{orth}}\) discourages redundant shared and private decoder directions.

\paragraph{Forecast-consistency fine-tuning.}
The complete \textsc{TSAE} training protocol then performs a short second-stage refinement through the frozen forecasting backbone. Let \(f\) denote the fixed forecasting map from the adapted hidden representation to the target forecast, and let \(\hat h\) be the \textsc{TSAE} reconstruction of \(h\). We optimize only \textsc{TSAE} parameters with
\begin{equation}
\label{eq:forecast-consistency}
L_{\mathrm{FT}}=L_{\mathrm{base}}+\lambda_{\mathrm{fc}}
\frac{\operatorname{MSE}(f(\hat h),f(h))}
{\operatorname{Var}(f(h))}.
\end{equation}
This forecast-consistency term encourages the reconstructed hidden representation to preserve the frozen backbone's output while retaining the first-stage reconstruction, sparsity, auxiliary, and temporal objectives. In the PatchTST headline runs, \(\lambda_{\mathrm{fc}}=0.1\), and this stage uses two epochs with at most 80 training batches per epoch. It is a component of the complete \textsc{TSAE} method; the forecasting backbone itself is never updated.

This objective is intended as an inductive bias rather than a guarantee of unique factor recovery. It encourages reusable temporal motifs to be represented by shared gated features, variable-conditioned residual structure by compact private features, and similar adjacent ordered temporal tokens by stable sparse codes. We therefore evaluate \textsc{TSAE} as a structured alternative to flat shared SAEs, rather than claiming uniform dominance on every metric.

\section{\textsc{TSEval}: Time-Series SAE Evaluation Protocol}

\textsc{TSEval} is a companion protocol for auditing the sparse features learned by \textsc{TSAE} and baseline SAE variants. Its role is deliberately narrower than the method section: \textsc{TSAE} defines how sparse features are learned, while \textsc{TSEval} defines which interpretability claims those features support. The protocol evaluates six axes, each targeting a failure mode that reconstruction-only SAE evaluation cannot rule out: reconstruction fidelity, feature coherence, physical calibration, forecast intervention sensitivity, hidden-space controls, and input-level baselines. Table~\ref{tab:checklist} summarizes the diagnostic question associated with each axis.

\begin{table}[!htb]
\centering
\footnotesize
\begin{tabular}{p{0.30\linewidth}p{0.57\linewidth}}
\toprule
\textbf{Diagnostic} & \textbf{Question} \\
\midrule
Fidelity & Does the sparse reconstruction preserve the original hidden state? \\
Coherence & Are activations concentrated in coherent variable and temporal contexts? \\
Calibration & Do feature activations track independently measurable signal motifs? \\
Forecast sensitivity & How sensitive is the forecast to a specified sparse-feature intervention? \\
Hidden control & Is the effect specific to the SAE decoder path rather than a generic hidden-space perturbation? \\
Input baselines & Do hidden sparse features provide information beyond saliency and occlusion? \\
\bottomrule
\end{tabular}
\caption{The \textsc{TSEval} diagnostic checklist. Each axis tests a claim that reconstruction-only evaluation cannot adjudicate. In \textsc{TSEval}, a feature is interpreted only with respect to the claims supported by the corresponding diagnostics.}
\label{tab:checklist}
\end{table}

A complete \textsc{TSEval} run takes a frozen forecasting backbone, hidden tokens with metadata, an SAE configuration, and raw segments for physical-motif computation. It then adapts hidden states to a common token contract; trains or loads the sparse dictionary; scores fidelity, coherence, and physical calibration; measures forecast sensitivity through decoder-mediated feature ablations and hidden-direct controls; and compares hidden-feature signals with saliency and occlusion baselines. Coherence is operationalized by semantic/channel concentration statistics, including purity scores for compatibility with prior SAE analyses.

\paragraph{Diagnostic definitions.}
Fidelity is the SAE reconstruction error on frozen backbone hidden tokens. Calibration selects the strongest of eight raw-segment motifs and reports standardized-fit residuals as calibration error. Forecast sensitivity is measured by decoder-mediated ablation: with \(z=E(h(x))\), \(\hat h=D(z)\), and \(\hat h^{(-i)}=D(z_{\setminus i})\),
\begin{equation}
\label{eq:feature-intervention-sensitivity}
\mathrm{FIS}_i =
\mathbb{E}_{(x,y)}
\left[
\ell(f(\hat h^{(-i)}),y)-\ell(f(\hat h),y)
\right].
\end{equation}
The hidden-direct control subtracts the matched direction \(z_i d_i\) from the original hidden state to test whether the forecast change is specific to the SAE decoder path. FIS is therefore an intervention-sensitivity diagnostic, not a causal effect estimator.

\paragraph{Capability-graded execution.}
Adapters normalize backbone-specific representations into \(H\in\mathbb{R}^{B\times C\times D\times P}\) when the corresponding axes are exposed, and record capability flags for decoding, temporal order, hidden controls, decoder marks, and cycle indices. Diagnostics are enabled only when the required token axes and decoding capabilities are available; otherwise the axis is marked unsupported rather than inferred from artificial metadata. The released toolkit covers 18 registered forecasting models through five adapter families, but smoke tests validate plug-in coverage and reproducibility, not scientific performance.

\paragraph{Why multiple axes are necessary.}
The axes are intentionally non-equivalent. Fidelity does not distinguish a meaningful shared temporal motif from uncontrolled cross-variable mixing. Coherence can reflect contextual concentration without tracking any physical motif. Calibration can align a feature with slope or volatility even when the forecasting head is insensitive to that decoder direction. \textsc{TSEval} therefore treats reconstruction as a precondition, physical correlation as semantic alignment rather than forecast causality, and FIS as forecast sensitivity under a specified intervention. Any feature-level interpretation should be restricted to the claims supported by its corresponding diagnostics.

\section{Experiments}

\input{sections/experiments_v10}

\section{Limitations}

The main evidence is strongest for PatchTST on the three ETT datasets.
This is a deliberate choice: it lets us compare sparse-feature methods under a
controlled hidden representation and a dataset-specific active-feature budget.
It also leaves open how far the same design transfers to other forecasting
architectures and higher-dimensional datasets; those questions require a
separate, fully matched evaluation.

The diagnostic metrics should also be read separately. \textsc{TSAE} is most
consistent on reconstruction and NFRE, but it is not uniformly best on every
interpretability proxy. JumpReLU or TopK can be stronger on selectivity,
physical correlation, or calibration error, depending on the dataset. Physical
motifs are hand-designed shape statistics, so they may miss domain-specific
semantics. FIS-style interventions measure sensitivity under a specified SAE
decoder intervention; they are not exhaustive causal attribution. Finally,
conditional temporal continuity is meaningful only when the adapter exposes
ordered temporal tokens. For non-patch or unordered token representations, the
temporal graph is empty and the term should not be overinterpreted.

\section{Conclusion}

We introduced \textsc{TSAE}, a structured sparse autoencoder for forecasting
representations. Its shared/private dictionaries separate reusable temporal
motifs from variable-conditioned residual structure, its gated parameterization
protects activation magnitudes from sparsity-induced shrinkage, and its
conditional temporal term discourages feature flicker on similar adjacent
patches. Under an active-matched protocol on ETTh1, ETTh2, and ETTm1,
\textsc{TSAE} achieves the lowest reconstruction error and NFRE among the five
compared methods on all three datasets.

The result is not a blanket dominance claim. The stronger conclusion is that
complete \textsc{TSAE} improves fidelity to forecast-relevant hidden
information while exposing real tradeoffs with selectivity, physical
correlation, and calibration. The reported ablations are consistent with
routing improving temporal stability, gating avoiding low realized activity,
and temporal continuity reducing patch-token flicker. These findings support
\textsc{TSAE} as a practical, structure-aware alternative to flat SAEs for
time-series forecasting interpretation.

\clearpage
\section*{AI Use Statement}
{\small
The authors used large language models (LLMs) as a research aid in preparing this paper. LLM assistance covered writing and polishing portions of the text, writing and debugging analysis code, brainstorming experimental designs and interpreting experimental results, and discovering relevant literature. All LLM-assisted output was verified and curated by the authors, who take full responsibility for the content of this paper. LLMs were not used to fabricate results and are not listed as authors.
}

\section*{Reproducibility Statement}
{\small
All headline comparisons use three independently initialized SAE runs; run-level values, operating points, and audit trails are reported in the appendices. Code, configurations, and the data-processing pipeline will be released publicly with a research-use license upon publication. Appendix~\ref{sec:appendix-experiments} documents the evaluation protocol and Appendix~\ref{sec:appendix-repro} the computational environment.
}

\bibliographystyle{iclr2027_conference}
\bibliography{references}

\clearpage
\appendix

\section{Case Study: What Does a \textsc{TSAE} Feature Represent?}
\label{sec:appendix-case-study}
\input{sections/appendix_case_study}

\input{sections/related_work}

\section{Additional Experimental Details}
\label{sec:appendix-experiments}

\input{sections/appendix_v10_experiments}

\FloatBarrier
\section*{Reproducibility Checklist}
\small

\paragraph{General paper information.}
\begin{itemize}
\item Includes a conceptual outline and/or pseudocode description of AI methods introduced: \textbf{yes}. Section ``\textsc{TSAE}: Structured Sparse Autoencoder'' defines the shared/private dictionaries, gated encoder, temporal regularizer, and training objective; Section ``\textsc{TSEval}: Time-Series SAE Evaluation Protocol'' gives the five-stage diagnostic procedure and the six diagnostic axes.
\item Clearly delineates opinions, hypotheses, and speculation from objective facts and results: \textbf{yes}. Interpretation rules, evidence hierarchy, limitations, and the supporting-evidence labels separate measured results from diagnostic recommendations.
\item Provides pedagogical references for readers to gain background necessary to replicate the paper: \textbf{yes}. Related work covers SAE evaluation, concept testing, time-series explanation, and forecasting architectures.
\end{itemize}

\paragraph{Theoretical contributions.}
Does this paper make theoretical contributions? \textbf{no}. The paper is primarily empirical and methodological. The argument that the six \textsc{TSEval} diagnostic axes are non-redundant is presented as a conceptual motivation within the protocol section rather than as formal propositions.
\begin{itemize}
\item Formal assumptions and statements are provided: \textbf{NA}.
\item Proof sketches or justifications are provided: \textbf{NA}.
\item Theoretical claims are connected to experiments: \textbf{NA}.
\item Code for theoretical claims: \textbf{NA}.
\end{itemize}

\paragraph{Datasets.}
Does this paper rely on one or more datasets? \textbf{yes}.
\begin{itemize}
\item Motivation for selected datasets: \textbf{yes}. Section ``Setup and Evidence Hierarchy'' motivates ETT and Weather as public long-horizon forecasting benchmarks with different variable counts and sampling frequencies.
\item Novel datasets introduced: \textbf{NA}. The paper introduces no new dataset.
\item Existing datasets are accompanied by appropriate citations: \textbf{yes}. The paper cites prior forecasting work using ETT/Weather, and the artifact README lists dataset sources.
\item Existing datasets are publicly available: \textbf{yes}. The artifact README provides public dataset locations.
\item Non-public datasets: \textbf{NA}. No private dataset is used.
\end{itemize}

\paragraph{Computational experiments.}
Does this paper include computational experiments? \textbf{yes}.
\begin{itemize}
\item Number and range of hyperparameter values tried during development, plus final selection criterion: \textbf{yes}. The main text lists headline settings, and the artifact reproducibility appendix records final settings, development-run families, and the selection rule used to choose the reported matrix.
\item Pre-processing code is included in the appendix/artifact: \textbf{yes}. Data loading, hidden-cache construction, input baselines, and aggregation entry points are included in the released scripts.
\item Source code required for conducting and analyzing experiments is included in a code appendix/artifact: \textbf{yes}. The artifact contains benchmark aggregation, all-model adapter smoke tests, and validation-sweep entry points; the repository contains the analysis implementation.
\item Source code will be publicly available upon publication with a research-use license: \textbf{yes}. The intended release is the submitted repository/artifact package.
\item Source code implementing new methods has comments detailing implementation links to the paper: \textbf{yes}. The adapter registry and SAE-analysis entry point include paper-to-code mapping comments, and the reproducibility appendix maps claims to scripts and artifacts.
\item Randomness and seed handling are described sufficiently to replicate results: \textbf{yes}. Released scripts expose separate backbone and SAE initialization variables, and the appendix reports every run-level value.
\item Computing infrastructure is specified: \textbf{yes}. The appendix records the final GPU and software environment, and the artifact contains the environment-capture output.
\item Evaluation metrics are formally described and motivated: \textbf{yes}. Reconstruction, semantic/channel purity, physical calibration, FIS, hidden-direct controls, and input baselines are defined in the protocol section.
\item Number of algorithm runs used to compute each reported result is stated: \textbf{yes}. Headline results use three independent SAE runs; single-run evidence is explicitly marked as supporting.
\item Analysis goes beyond single-dimensional summaries: \textbf{yes}. The paper reports mean and standard deviation, paired run directions, multi-axis diagnostics, and failure-mode analyses.
\item Significance of improvements/decreases is judged with appropriate statistical checks: \textbf{NA}. With three runs, we make no inferential significance claim; uncertainty is reported descriptively and all run-level values are exposed.
\item Final hyperparameters used for each model/algorithm are listed: \textbf{yes}. Main-text settings, scripts, per-run configuration files, and the artifact reproducibility appendix specify the headline configurations.
\end{itemize}

\end{document}

%% file: sections/experiments_v10.tex
\subsection{Experimental Setup}

\paragraph{Backbone, datasets, and baselines.}
We evaluate \textsc{TSAE} on frozen PatchTST patch-token representations. The
core benchmark compares \textsc{TSAE} with L1, Gated, TopK, and JumpReLU SAEs
on ETTh1, ETTh2, and ETTm1 under common data splits and a 2,048-feature
per-token encoding width. This matches per-token code width, not the total
routed dictionary parameter count. \textsc{TSAE}'s complete protocol includes
the forecast-consistency refinement in Section~\ref{sec:tsae-method}; baselines
retain their original SAE objectives, while all forecasting backbones remain
frozen. Weather is an external-validity extension. Validation data select
comparable operating points, and the PatchTST--ETTh1 validation sweep compares
fidelity against realized activity. Main-table results are three-seed means
$\pm$ standard deviations; iTransformer and DLinear are transfer checks.

\paragraph{Evaluation protocol.}
For the PatchTST--ETTh1 activity--fidelity analysis, we sweep each method's
sparsity controls on validation data over a broad range of realized activity
levels (roughly $10^2$--$10^3$ active features per token where supported by the
method). Fidelity--sparsity tradeoffs are compared using measured activity,
rather than raw hyperparameter values. For the formal three-seed PatchTST
comparison, we fix a validation-selected configuration within a prespecified
comparable activity band before retraining from scratch on the training split
and evaluating each seed once on the test split. TopK uses an exact integer
activity budget. The full selection rule and selected operating points are
reported in the appendix.

The PatchTST core evaluation reports reconstruction MSE, normalized
forecast-reconstruction error (NFRE), mean active features, dead-feature ratio,
ordered-code transition rate, selectivity AUROC, physical motif correlation,
and calibration error. Reconstruction MSE and NFRE quantify forecast-relevant
fidelity; mean activity and dead-feature ratio measure sparsity and dictionary
utilization; the remaining diagnostics assess temporal stability, selectivity,
physical alignment, and calibration. Cross-backbone transfer checks report the
subset of metrics that is comparably defined for their representation geometry.

\subsection{Main Results}

Our empirical evidence is organized around a controlled PatchTST comparison
and two complementary analyses. The headline evaluation fixes the forecasting
representation and compares five SAE methods on ETTh1, ETTh2, and ETTm1 under
the validation-selected, three-seed test protocol; Weather is reported under
the same protocol as an external-validity extension. We then test whether the
fidelity pattern persists across realized activity levels with a validation-only
curve,
and isolate the contributions of routing, gating, and temporal continuity by
ablation. The appendix records validation selection, full configurations, and
the iTransformer and DLinear extensions; these extensions are supporting
evidence rather than replacements for the controlled comparison.

\input{sections/main_results_v10_table}

\paragraph{Controlled PatchTST comparison.}
Table~\ref{tab:v10-main-results} reports the complete five-method comparison
across four datasets. \textsc{TSAE} has the lowest reconstruction MSE and NFRE
on all four. On the three ETT datasets, its NFRE reductions relative to the
next-best mean are 4.3\% (ETTh1), 26.4\% (ETTh2), and 9.1\% (ETTm1); on
Weather the reduction is 11.3\%. Reconstruction MSE follows the same ordering,
with reductions of 1.0--24.4\%. Operating points are selected under the
prespecified validation active-feature window. For the three ETT datasets,
every reported test mean lies in \([192,320]\). Weather \textsc{TSAE}, Gated, TopK, and
JumpReLU also remain in-window; the Weather L1 test mean is slightly below the
boundary (187.8) despite its validation-selected point being in-window. The
active-feature protocol controls realized per-token activity; total routed
dictionary parameter counts are reported separately in the appendix.

\begin{figure}[!htb]
\centering
\includegraphics[width=0.98\columnwidth]{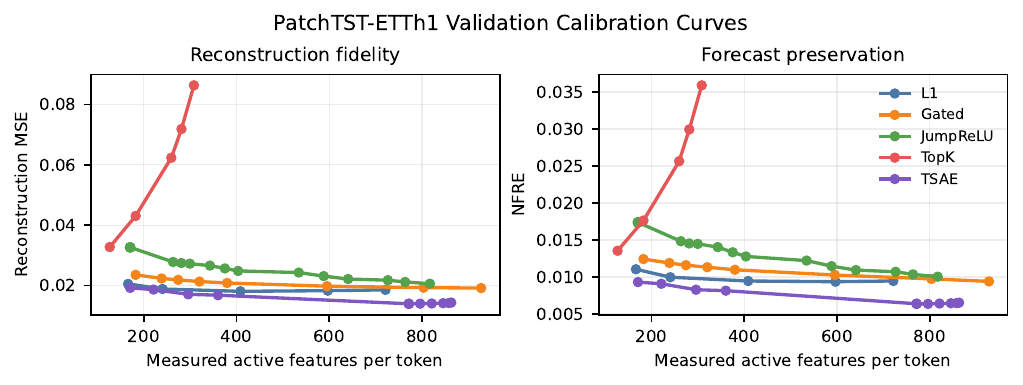}
\caption{PatchTST--ETTh1 validation activity--fidelity curves. Each point is a
validation calibration run; the horizontal axis reports measured active
features per token. Table~\ref{tab:v10-main-results} reports separate
test-split metrics.}
\label{fig:v10-active-fidelity}
\end{figure}

\paragraph{Activity--fidelity analysis.}
Table~\ref{tab:v10-main-results} compares methods at validation-selected
operating points, but does not show whether the fidelity advantage is confined
to one realized activity level. We therefore sweep SAE sparsity controls on
the PatchTST--ETTh1 validation split and plot fidelity against measured active
features per token in Figure~\ref{fig:v10-active-fidelity}. This validation
analysis complements, rather than replaces, the test-split comparison in
Table~\ref{tab:v10-main-results}.

In the sampled overlapping activity range, \textsc{TSAE} has lower
reconstruction and forecast-reconstruction error than the displayed baselines.
Its Table~\ref{tab:v10-main-results} fidelity advantage is therefore not tied
to a single selected activity level. Because Figure~\ref{fig:v10-active-fidelity}
is a single-dataset validation analysis, we treat it as supporting robustness
evidence rather than an additional headline result.

To test whether the fidelity advantage transfers beyond PatchTST, we ran the
same active-matched protocol on frozen iTransformer--ETTh1 and
DLinear--ETTh1 representations. Table~\ref{tab:cross-backbone} summarizes
the results with the PatchTST comparison for reference.

\begin{table}[!htb]
\centering
\footnotesize
\setlength{\tabcolsep}{3.5pt}
\begin{tabular}{@{}llrrrr@{}}
\toprule
Backbone & Method & Active & Recon.\ $\downarrow$ & NFRE $\downarrow$ \\
\midrule
PatchTST    & TSAE  & 231.4$\pm$2.7  & \textbf{13.44$\pm$0.57} & \textbf{5.10$\pm$0.25} \\
            & Gated & 244.4$\pm$7.4  & 16.39$\pm$0.75 & 6.39$\pm$0.22 \\
            & L1    & 252.8$\pm$10.3 & \underline{13.59$\pm$0.45} & \underline{5.33$\pm$0.27} \\
            & TopK  & 264.9$\pm$5.7  & 30.54$\pm$2.00 & 10.84$\pm$0.75 \\
            & JumpReLU & 269.8$\pm$11.4 & 20.69$\pm$0.88 & 8.09$\pm$0.37 \\
\midrule
iTransformer & TSAE  & 253.1$\pm$2.5  & \textbf{14.48$\pm$0.23} & \textbf{3.01$\pm$0.06} \\
             & Gated & 315.7$\pm$1.1  & \underline{17.29$\pm$0.13} & \underline{3.20$\pm$0.09} \\
             & L1    & 282.3$\pm$1.0  & 18.12$\pm$0.08 & 4.14$\pm$0.20 \\
             & TopK  & 211.8$\pm$0.3  & 30.10$\pm$0.14 & 5.24$\pm$0.15 \\
             & JumpReLU & 268.6$\pm$3.4 & 18.84$\pm$0.11 & 3.44$\pm$0.05 \\
\midrule
DLinear      & TSAE  & 215.0$\pm$10.2 & 9.09$\pm$0.16  & 7.91$\pm$0.23 \\
             & Gated & 314.1$\pm$1.1  & \textbf{7.71$\pm$0.09}  & \textbf{6.88$\pm$0.14} \\
             & L1    & 250.4$\pm$1.2  & 9.17$\pm$0.09  & 8.74$\pm$0.14 \\
             & TopK  & 256.0$\pm$0.0  & 9.90$\pm$0.21 & 8.27$\pm$0.22 \\
             & JumpReLU & 266.0$\pm$3.7 & \underline{7.99$\pm$0.12} & \underline{7.02$\pm$0.16} \\
\bottomrule
\end{tabular}
\caption{Cross-backbone active-matched comparison (ETTh1, three-seed means).
Bold and underline mark the best and second-best means, respectively, within
each backbone and metric. Recon.\ and NFRE are
$\times10^{3}$. Full tables and validation sweeps are in
Appendix~\ref{sec:appendix-cross-backbone}.}
\label{tab:cross-backbone}
\end{table}

\paragraph{Cross-backbone diagnostic.}
On PatchTST and iTransformer, \textsc{TSAE} yields the lowest reconstruction
and NFRE, consistent with the primary result. On DLinear, Gated SAE has the
best reconstruction and NFRE, indicating that the fidelity advantage does not
automatically transfer to every architecture. These cross-backbone comparisons
should be interpreted as supporting evidence rather than as replacements for
the controlled PatchTST headline evaluation.

This fidelity advantage does not imply uniform dominance on every diagnostic.
L1 has the lowest dead-feature ratio and calibration error on several datasets;
TopK and JumpReLU lead selectivity or physical correlation in some settings.
\textsc{TSAE} instead makes a narrower claim: under a controlled activity
budget, its structured code preserves forecast-relevant hidden information
best and yields the lowest transition rate in the four reported PatchTST
comparisons. The appendix
contains the selected settings and validation-sweep record.

\subsection{Ablation Study}

The ablation examines the behavior of \textsc{TSAE} components on the PatchTST
backbone with ETTh1, ETTh2, and ETTm1. We compare the full model with
variants that remove private routing, gated parameterization, or the
temporal-continuity term. The three-seed results in
Table~\ref{tab:v10-ablation} summarize changes in activity, fidelity,
dictionary utilization, and temporal stability. They are component
diagnostics, rather than capacity- or activity-matched causal estimates.

\begin{table}[!htb]
\centering
\footnotesize
\setlength{\tabcolsep}{2.1pt}
\begin{tabular}{@{}llrrrrr@{}}
\toprule
Data & Variant & Active & Recon. $\downarrow$ & NFRE $\downarrow$ & Dead & Trans. \\
\midrule
ETTh1 & Full      & 268 & \textbf{11.7} & \textbf{4.9} & 6.3  & 0.086 \\
      & No routing & 252 & 13.4 & 5.5 & 0.0  & 0.145 \\
      & No gating  & 108 & 18.1 & 7.9 & 0.0  & 0.083 \\
      & No temporal& 279 & 12.3 & 5.1 & 6.3  & 0.091 \\
\midrule
ETTh2 & Full      & 203 & \textbf{2.2} & \textbf{3.4} & 10.9 & 0.036 \\
      & No routing & 178 & 2.4 & 3.6 & 13.1 & 0.050 \\
      & No gating  & 46  & 3.7 & 5.7 & 23.3 & 0.027 \\
      & No temporal& 234 & 2.8 & 4.2 & 13.3 & 0.038 \\
\midrule
ETTm1 & Full      & 218 & \textbf{9.1} & \textbf{5.2} & 3.8  & 0.049 \\
      & No routing & 192 & 9.8 & 5.7 & 0.1  & 0.071 \\
      & No gating  & 79  & 14.1 & 8.5 & 0.0  & 0.047 \\
      & No temporal& 261 & 10.3 & 5.8 & 6.2  & 0.052 \\
\bottomrule
\end{tabular}
\caption{PatchTST component ablation (three-seed means). Bold marks
the lowest reconstruction and NFRE values within each dataset. Dead-feature
ratio and transition rate are reported as diagnostic tradeoffs. Recon.\ and
NFRE use the report scale ($\times 10^{3}$).}
\label{tab:v10-ablation}
\end{table}


\paragraph{Component contributions.}
In the reported configurations, removing routing raises transition rate by
39--69\% across the three datasets (ETTh2: 0.036$\rightarrow$0.050;
ETTm1: 0.049$\rightarrow$0.071; ETTh1: 0.086$\rightarrow$0.145). Removing
gating reduces realized activity from 203--268 to 46--108 and coincides with
worse reconstruction, so it does not isolate a fidelity effect at matched
activity. The temporal term has a smaller anti-flicker effect: its
validation-selected coefficient is $\lambda_{\mathrm{temp}}=0.5$, whereas
$\lambda_{\mathrm{temp}}=0.001$ is indistinguishable from zero in the
sensitivity diagnostic.

\paragraph{Mechanism.}
These directions are consistent with the structural roles in
Section~\ref{sec:tsae-method}: routing allocates shared motifs and
variable-conditioned features separately, gating separates detection from
magnitude, and temporal continuity discourages switching on similar adjacent
patches. The ablation alone does not identify these mechanisms apart from
changes in total capacity or realized activity.

%% file: sections/main_results_v10_table.tex
\begin{table}[!htb]
\centering
\scriptsize
\setlength{\tabcolsep}{1.2pt}
\renewcommand{\arraystretch}{1.08}
\begin{tabular}{@{}llrrrrrrrr@{}}
\toprule
\multicolumn{2}{c}{Setting} & \multicolumn{1}{c}{Operating point} & \multicolumn{7}{c}{Evaluation metrics} \\
\cmidrule(r){1-2}\cmidrule(lr){3-3}\cmidrule(l){4-10}
Data & Method & Active & Recon. $\downarrow$ & NFRE $\downarrow$ & Dead $\downarrow$ & Trans. $\downarrow$ & Sel. $\uparrow$ & Phys. $\uparrow$ & Calib. $\downarrow$ \\
\midrule
ETTh1 & TSAE & 231.4$\pm$2.7  & \textbf{13.44$\pm$0.57} & \textbf{5.10$\pm$0.25} & 0.080$\pm$0.025 & \textbf{0.073$\pm$0.005} & \textbf{0.639$\pm$0.007} & 0.541$\pm$0.106 & 0.624$\pm$0.049 \\
 & L1 & 252.8$\pm$10.3 & \underline{13.59$\pm$0.45} & \underline{5.33$\pm$0.27} & \textbf{0.000$\pm$0.000} & 0.181$\pm$0.006 & 0.570$\pm$0.023 & \textbf{0.653$\pm$0.047} & \textbf{0.567$\pm$0.022} \\
 & Gated & 244.4$\pm$7.4  & 16.39$\pm$0.75 & 6.39$\pm$0.22 & \underline{0.000$\pm$0.000} & 0.144$\pm$0.005 & 0.573$\pm$0.020 & 0.523$\pm$0.060 & 0.681$\pm$0.024 \\
 & TopK & 264.9$\pm$5.7  & 30.54$\pm$2.00 & 10.84$\pm$0.75 & 0.001$\pm$0.001 & \underline{0.143$\pm$0.002} & 0.592$\pm$0.007 & \underline{0.622$\pm$0.047} & \underline{0.587$\pm$0.043} \\
 & JumpReLU & 269.8$\pm$11.4 & 20.69$\pm$0.88 & 8.09$\pm$0.37 & \textbf{0.000$\pm$0.000} & 0.156$\pm$0.007 & \underline{0.612$\pm$0.066} & 0.587$\pm$0.017 & 0.657$\pm$0.002 \\
\midrule
ETTh2 & TSAE & 250.9$\pm$13.6 & \textbf{6.04$\pm$0.29} & \textbf{2.70$\pm$0.14} & 0.043$\pm$0.014 & \textbf{0.050$\pm$0.002} & 0.327$\pm$0.001 & 0.571$\pm$0.066 & 0.595$\pm$0.027 \\
 & L1 & 195.2$\pm$11.4 & \underline{7.99$\pm$0.18} & \underline{3.66$\pm$0.23} & \textbf{0.025$\pm$0.016} & 0.106$\pm$0.009 & \textbf{0.335$\pm$0.005} & 0.600$\pm$0.087 & \underline{0.581$\pm$0.036} \\
 & Gated & 212.7$\pm$6.0  & 8.19$\pm$0.46 & 3.76$\pm$0.20 & 0.103$\pm$0.026 & \underline{0.055$\pm$0.003} & 0.327$\pm$0.005 & \underline{0.604$\pm$0.040} & 0.593$\pm$0.016 \\
 & TopK & 218.3$\pm$1.0  & 9.88$\pm$0.37 & 4.56$\pm$0.20 & 0.143$\pm$0.016 & 0.063$\pm$0.003 & 0.329$\pm$0.004 & 0.498$\pm$0.047 & 0.671$\pm$0.027 \\
 & JumpReLU & 311.0$\pm$39.9 & 10.86$\pm$0.74 & 4.79$\pm$0.27 & \underline{0.041$\pm$0.015} & 0.064$\pm$0.007 & \underline{0.333$\pm$0.006} & \textbf{0.752$\pm$0.048} & \textbf{0.520$\pm$0.025} \\
\midrule
ETTm1 & TSAE & 204.9$\pm$11.5 & \textbf{10.35$\pm$0.48} & \textbf{5.00$\pm$0.28} & 0.031$\pm$0.004 & \textbf{0.047$\pm$0.002} & 0.615$\pm$0.032 & \underline{0.571$\pm$0.068} & 0.673$\pm$0.036 \\
 & L1 & 258.0$\pm$11.9 & \underline{11.44$\pm$0.65} & \underline{5.50$\pm$0.25} & \textbf{0.000$\pm$0.000} & 0.135$\pm$0.005 & 0.620$\pm$0.038 & 0.565$\pm$0.127 & \textbf{0.608$\pm$0.055} \\
 & Gated & 246.9$\pm$5.8  & 13.49$\pm$0.80 & 6.58$\pm$0.41 & \underline{0.000$\pm$0.000} & 0.089$\pm$0.002 & 0.617$\pm$0.007 & 0.548$\pm$0.033 & 0.677$\pm$0.022 \\
 & TopK & 223.7$\pm$2.4  & 19.92$\pm$0.59 & 8.02$\pm$0.16 & 0.006$\pm$0.001 & \underline{0.083$\pm$0.001} & \textbf{0.638$\pm$0.013} & \textbf{0.598$\pm$0.050} & \underline{0.623$\pm$0.025} \\
 & JumpReLU & 269.9$\pm$12.8 & 17.09$\pm$0.97 & 8.63$\pm$0.27 & \textbf{0.000$\pm$0.000} & 0.092$\pm$0.004 & \underline{0.630$\pm$0.019} & 0.568$\pm$0.108 & 0.669$\pm$0.024 \\
\midrule
Weather & TSAE      & 216.7$\pm$3.2  & \textbf{1.70$\pm$0.06} & \textbf{2.52$\pm$0.22} & 0.195$\pm$0.006 & \textbf{0.019$\pm$0.000} & \underline{0.574$\pm$0.016} & 0.310$\pm$0.016 & 0.768$\pm$0.006 \\
        & L1         & 187.8$\pm$16.3 & \underline{1.85$\pm$0.15} & 3.37$\pm$0.22 & \textbf{0.000$\pm$0.000} & 0.055$\pm$0.003 & 0.557$\pm$0.005 & 0.310$\pm$0.005 & \textbf{0.725$\pm$0.006} \\
        & Gated      & 285.1$\pm$8.8  & 2.10$\pm$0.15 & \underline{2.84$\pm$0.13} & \underline{0.000$\pm$0.000} & 0.046$\pm$0.002 & 0.571$\pm$0.021 & \underline{0.319$\pm$0.014} & 0.765$\pm$0.007 \\
        & TopK       & 261.1$\pm$14.4 & 3.84$\pm$0.15 & 4.64$\pm$0.35 & 0.073$\pm$0.004 & 0.038$\pm$0.002 & 0.533$\pm$0.018 & 0.312$\pm$0.010 & \underline{0.755$\pm$0.004} \\
        & JumpReLU   & 236.0$\pm$0.7  & 4.49$\pm$0.53 & 6.21$\pm$0.84 & 0.001$\pm$0.001 & \underline{0.031$\pm$0.002} & \textbf{0.577$\pm$0.016} & \textbf{0.321$\pm$0.006} & 0.765$\pm$0.008 \\
\bottomrule
\end{tabular}
\caption{Validation-selected PatchTST results on four datasets. Each entry is the mean$\pm$standard deviation over three independently trained SAE seeds. Operating points are selected only on validation data to target 256 active features per token (admissible range: 192--320); all reported metrics are test-split measurements. Bold and underline mark the best and second-best mean, respectively, within a dataset and metric. Recon. and NFRE are multiplied by $10^3$.}
\label{tab:v10-main-results}
\end{table}

%% file: sections/appendix_case_study.tex
Aggregate diagnostics establish fidelity and code-level behavior, but they do
not by themselves show what an individual atom represents. We therefore
conduct an auditable case study using the formal \textsc{TSAE} PatchTST
checkpoints and test hidden-state caches. The protocol treats a private atom
as a variable-specific pair (variable index, local atom index), rather than
merging the same local coordinate across private dictionaries, and combines
four complementary checks: repeated raw-patch patterns, variable scope,
physical strength alignment, and forecast sensitivity.

\begin{figure}[!htb]
\centering
\includegraphics[width=\textwidth]{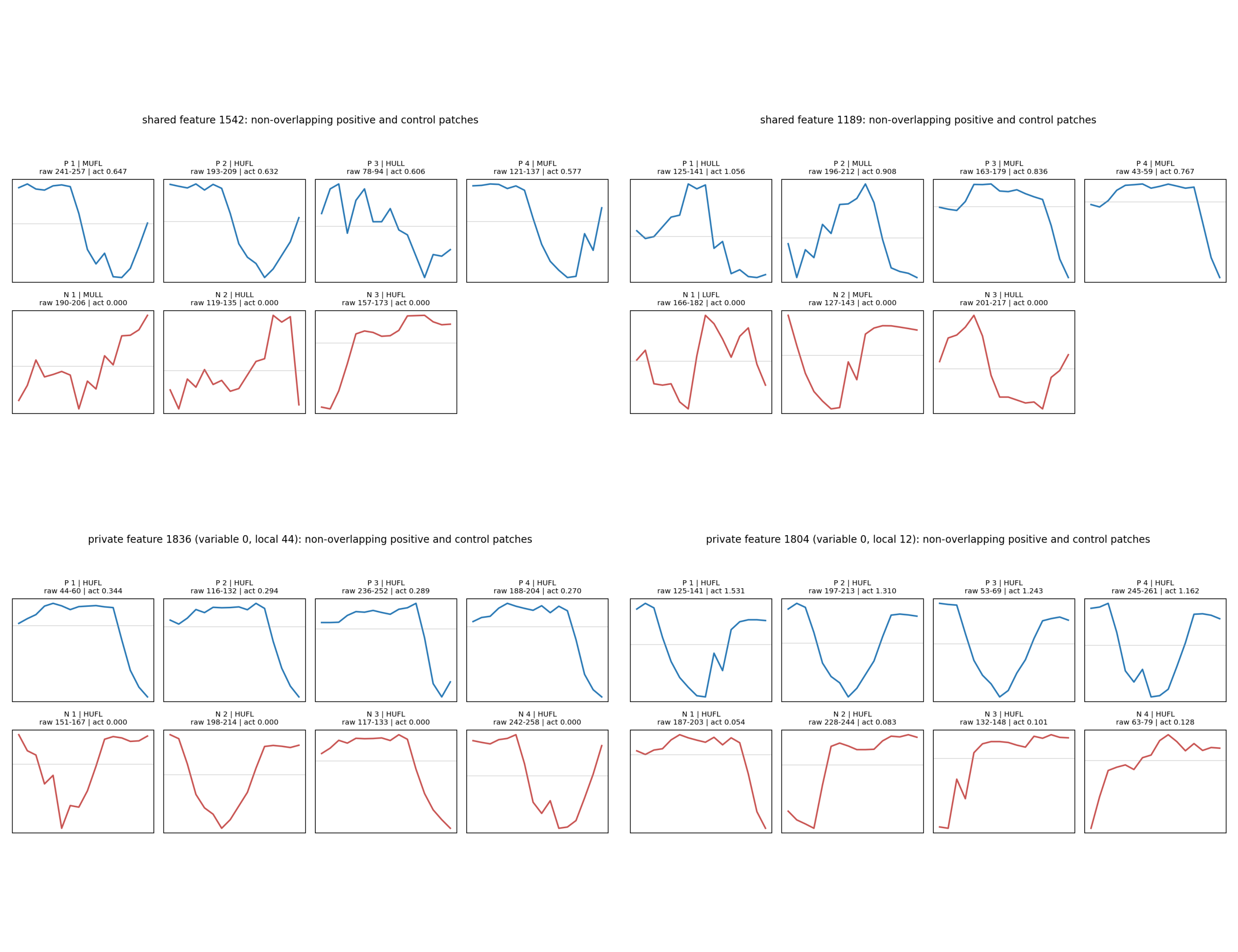}
\caption{Multi-feature case atlas from the formal PatchTST--ETTh1 run. Blue
curves are non-overlapping high-activation patches and red curves are
non-overlapping low-activation controls. The first row shows shared atoms that
recur across variables; the second row shows private atoms, each restricted to
its routed variable. The panel is illustrative; the full pre-recorded 12-atom
atlas and selection manifest are stored in the artifact. In-panel annotations
are small and are best viewed zoomed in the electronic PDF.}
\label{fig:v10-case-atlas}
\end{figure}

\paragraph{Repeated patterns and shared/private scope.}
Figure~\ref{fig:v10-case-atlas} displays two shared and two private atoms from
the formal ETTh1 run. For each atom, positive examples are chosen from
distinct raw intervals with a minimum separation of one patch; red controls
are also non-overlapping and have negligible activation. The shared atoms
recur across several variables, whereas each private atom is evaluated only on
its routed variable. Variable scope for private atoms is architectural, while
recurrence and physical alignment are checked empirically.

\begin{figure}[!htb]
\centering
\begin{minipage}[t]{0.48\textwidth}
\centering
\textbf{(a) Physical-strength calibration}\\[-2pt]
\includegraphics[width=\linewidth]{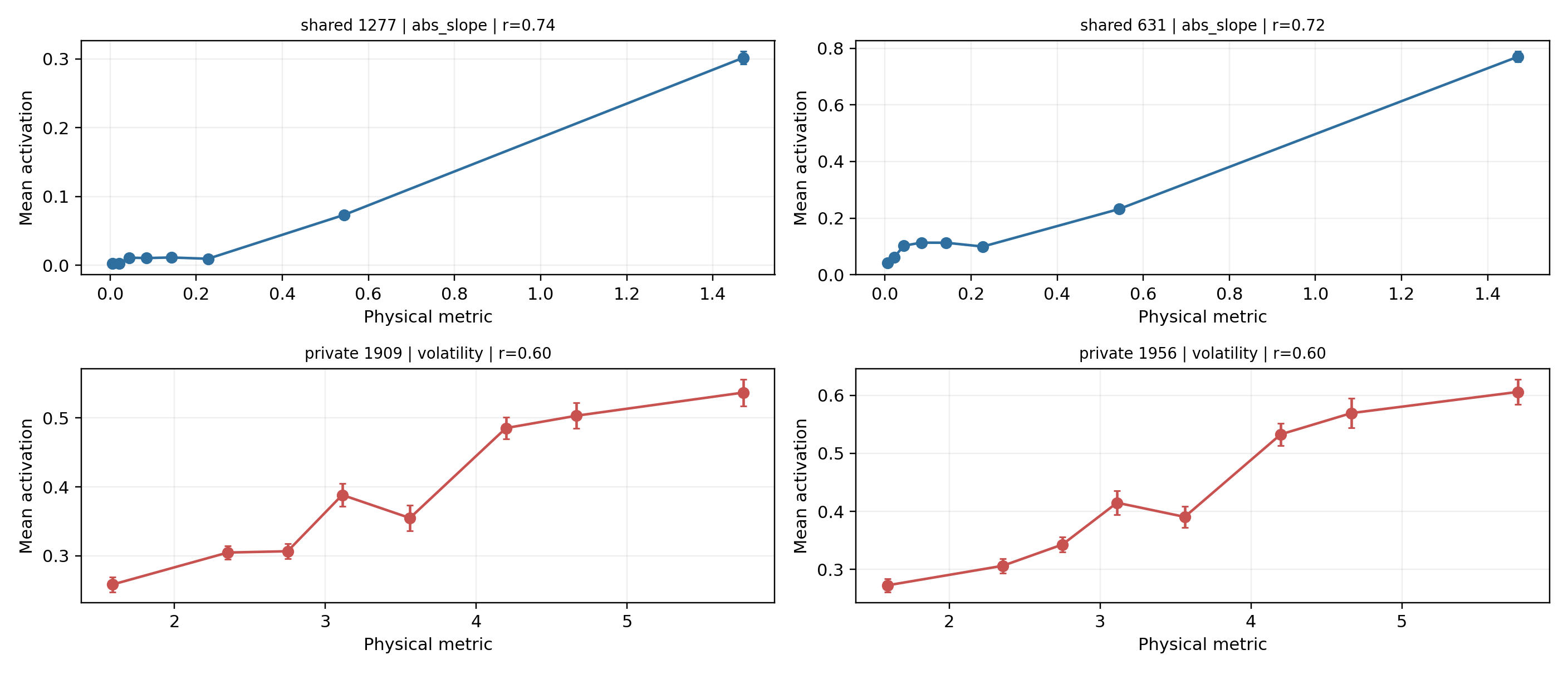}
\end{minipage}\hfill
\begin{minipage}[t]{0.48\textwidth}
\centering
\textbf{(b) Forecast intervention}\\[-2pt]
\includegraphics[width=\linewidth]{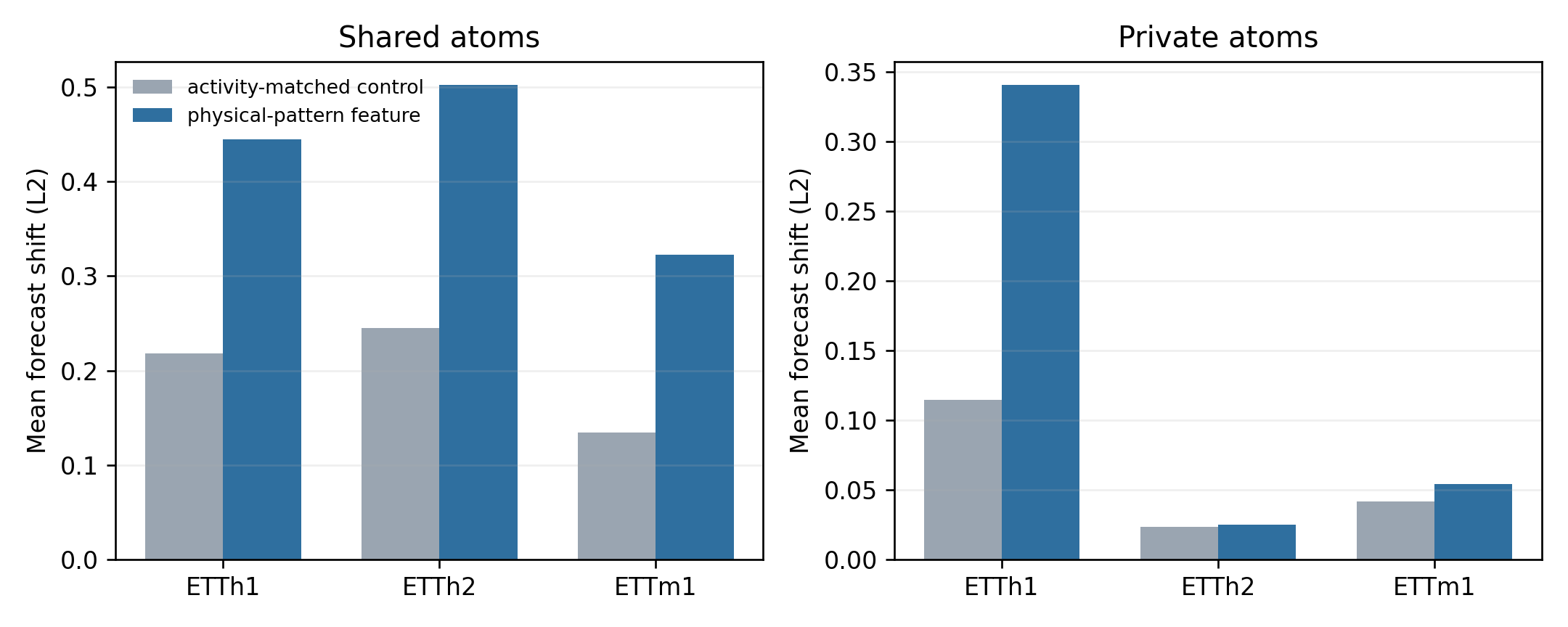}
\end{minipage}
\caption{Feature-level evidence beyond reconstruction. (a) ETTh1 activation
as a function of the associated raw physical metric; bars show 95\% confidence
intervals within equal-frequency bins. (b) Mean forecast shift after deleting
a selected physical-pattern feature (blue) or an activation-rate-matched,
lower-correlation control (gray), while the PatchTST backbone remains frozen.
The intervention measures sensitivity within the model pathway, not a causal
effect in the data-generating system.}
\label{fig:v10-case-evidence}
\end{figure}

\paragraph{Strength, continuity, and forecast relevance.}
Figure~\ref{fig:v10-case-evidence}(a) bins a raw physical metric into
equal-frequency intervals and plots the mean activation of four selected
atoms. The displayed atom--metric pairs were selected by their absolute
correlation from the pre-specified slope, volatility, range, curvature, and
spike descriptors, so the curves are descriptive illustrations rather than
held-out significance tests. The two displayed shared atoms track absolute
slope ($r=0.74$ and $0.72$), while the two private atoms track the volatility
of their routed variable ($r=0.60$ for both).

Temporal coherence is not inferred from these examples. Instead, after
retaining exactly the top 24 atoms per token for both \textsc{TSAE} and L1
SAE, the mean adjacent-patch Jaccard similarity is higher for \textsc{TSAE} on
all three datasets (Table~\ref{tab:appendix-case-temporal}). This
evaluation-only readout removes unequal support size as an explanation for the
comparison.

Figure~\ref{fig:v10-case-evidence}(b) tests forecast relevance by zeroing one
atom in the SAE reconstruction and decoding through the frozen forecasting
backbone. Each physically aligned atom is paired with a lower-correlation
control whose activation rate is within a factor of 1.4. Removing the selected
shared atoms produces larger mean forecast shifts than their controls on all
three datasets (ratios $2.04$, $2.05$, and $2.40$). Private-atom evidence is
strongest on ETTh1 (ratio $2.97$) and weaker on ETTh2/ETTm1 (ratios $1.06$
and $1.30$). Thus the case study supports forecast relevance for selected
atoms, but does not claim that every private atom has an equally large
downstream effect or that the intervention identifies external physical
causality.

\begin{table}[!htb]
\centering
\footnotesize
\begin{tabular}{@{}lrrr@{}}
\toprule
Dataset & L1 SAE Jaccard & \textsc{TSAE} Jaccard & Difference \\
\midrule
ETTh1 & 0.184 & 0.257 & +0.073 \\
ETTh2 & 0.456 & 0.676 & +0.220 \\
ETTm1 & 0.424 & 0.540 & +0.116 \\
\bottomrule
\end{tabular}
\caption{Adjacent-patch support stability at an identical Top-$24$ feature
budget per token. Values are descriptive means over adjacent test-patch pairs
for the seed-2024 formal checkpoints.}
\label{tab:appendix-case-temporal}
\end{table}

\paragraph{Audit trail.}
The case study uses the seed-2024 formal \textsc{TSAE} checkpoint for each ETT
dataset and the corresponding recorded test hidden-state cache; it is a
feature-level audit, not an additional three-seed method comparison. For the
atlas we select eight shared and four private atoms, require four positive
examples from distinct samples, and reject any pair of raw intervals that
overlap or are separated by fewer than 16 time steps; private controls are
drawn only from the owner variable. The artifact stores every selected
interval, rejection reason, activation, checkpoint hash, and cache hash.

For the temporal readout, we rank each method's features within each token and
retain the same top-$24$ support before computing adjacent-patch Jaccard
similarity. This is an evaluation-only normalization: it neither changes the
trained dictionaries nor estimates a new reconstruction metric, and
Table~\ref{tab:appendix-case-temporal} should not be compared directly to raw
transition rates, whose scale also depends on dictionary width.

For strength calibration, raw patches are assigned to equal-frequency bins of
the selected physical descriptor and we report the mean feature activation and
within-bin 95\% confidence interval. Shared atoms are assessed over all
variables; a private atom is assessed only over its routed variable, because
zero values outside that domain are structural rather than negative evidence.

For forecast interventions, a single selected atom is set to zero in the SAE
code, its reconstruction is decoded by the frozen PatchTST model, and the
forecast shift is measured relative to the unablated SAE reconstruction. Each
selected atom is paired with a lower physical-correlation control whose
activation rate differs by at most a factor of 1.4. The artifact contains 12
independent top-activation samples for each signal/control atom, per-sample
forecast shifts and MSE changes, pair identities, and manifests. This
establishes sensitivity along the frozen model pathway; it does not turn the
selected motif into a causal variable of the data-generating process.

%% file: sections/related_work.tex
\section{Related Work}
\label{sec:related-work}

\paragraph{Time-series explainability.}
Time-series explainability has traditionally focused on attributing predictions to input variables, timesteps, or subsequences. Saliency methods~\citep{sundararajan2017axiomatic}, LIME-style local approximations~\citep{ribeiro2016lime}, dynamic masks~\citep{crabbe2021dynamask}, Shapley-value methods such as TimeSHAP~\citep{bento2021timeshap}, occlusion, and counterfactual explanations identify which parts of an input sequence affect a forecast. Sanity checks for saliency~\citep{adebayo2018sanity} and attribution benchmarks~\citep{hooker2019benchmark} show that such explanations can be fragile. More importantly for this paper, input-level attribution answers a different question from hidden-feature interpretation: it can locate influential inputs, but it does not test whether an internal sparse feature is reconstructed, semantically coherent, physically calibrated, or forecast-sensitive.

\paragraph{Sparse autoencoders for representation interpretability.}
Sparse autoencoders have recently become a central tool for interpreting hidden representations in large language models. SAE work studies reconstruction--sparsity tradeoffs~\citep{gao2024scaling}, feature splitting and monosemanticity~\citep{bricken2023monosemanticity,bussmann2024matryoshka,lieberum2024does}, shrinkage and gated or JumpReLU variants~\citep{rajamanoharan2024improving,rajamanoharan2024jumprelu}, and downstream interventions. Related concept-based methods such as TCAV~\citep{kim2018tcav} and sparse probing~\citep{gurnee2023finding} test whether hidden directions align with human-interpretable concepts. These works motivate our use of sparse hidden features, but their standard evidence---reconstruction, sparsity, top activations, and text-domain interventions---does not directly address variable-indexed and physically scaled time-series representations.

\paragraph{SAE-based time-series explanations.}
Recent work has begun to adapt SAE-style explanations to time-series models,
but the role of forecasting hidden-state structure remains underexplored.
Forecasting work such as LaST separates latent seasonal and trend components to
improve accuracy~\citep{wang2022last}. It does not audit sparse hidden features
for cross-variable structure, temporal stability, or calibration.
TimeSAE is the closest time-series-specific sparse explanation framework to
our work~\citep{oublal2026timesae}: it learns sparse concepts to explain the
predictions of a black-box time-series model, emphasizing explanation
faithfulness under intervention, distribution shift, and compositional
consistency.

In contrast, \textsc{TSAE} learns structured sparse features from hidden
states extracted from a frozen forecasting backbone, rather than learning a
sparse explanation of a black-box predictor's input--prediction behavior. The
two perspectives are complementary: TimeSAE asks whether sparse concepts
faithfully explain a predictor's behavior, whereas \textsc{TSAE} asks how
internal forecasting features should be structured and audited for fidelity,
temporal stability, and physical calibration.

\paragraph{Evaluation of time-series sparse features.}
Prior SAE evaluation practices examine reconstruction--sparsity tradeoffs,
activation patterns, and downstream effects~\citep{gao2024scaling,bricken2023monosemanticity}.
These diagnostics do not, by themselves, directly assess variable structure,
temporal stability, or calibration to time-series signals. \textsc{TSAE}
specializes SAE design through shared/private dictionaries, magnitude-preserving
gates, and conditional temporal continuity. \textsc{TSEval} extends
sparse-feature evaluation with reconstruction, semantic/channel purity,
physical calibration, forecast intervention sensitivity, hidden controls, and
input-level baselines, so that each interpretability claim is tied to a
separate diagnostic rather than inferred from reconstruction alone.

%% file: sections/appendix_v10_experiments.tex
\subsection{Scope and Organization}

This appendix documents parameter selection, sensitivity analyses,
per-dataset active-matching diagnostics, cross-backbone extensions, and
reproducibility information. Section~\ref{sec:appendix-config} describes
how hyperparameters were chosen and reports the sweeps that justify those
choices. Section~\ref{sec:appendix-active-achievement} reports per-dataset
active-matched achievement. Section~\ref{sec:appendix-cross-backbone}
extends the comparison to iTransformer and DLinear backbones and reports
multi-backbone component ablations. Section~\ref{sec:appendix-repro}
provides reproducibility details. Appendix~\ref{sec:appendix-case-study}
records the feature-level case-study protocol and its activity-controlled
temporal readout.

\subsection{Configuration \& Sensitivity}
\label{sec:appendix-config}

\subsubsection*{Selection procedure}

For each dataset and each scale-sensitive method, we ran a validation sweep
over the sparsity-related parameter, recorded the mean active features per
token, and selected the setting whose validation activity was closest to 256.
Configurations with mean activity in \([192,320]\) were admissible; ties were
broken by validation reconstruction MSE. TopK used an exact activity budget
(\(k=256\) for the ETT datasets and \(k=320\) for Weather). The selected setting was
then fixed for three-seed test-split retraining.

\subsubsection*{Selected operating points}

\begin{table}[!htb]
\centering
\footnotesize
\setlength{\tabcolsep}{2.8pt}
\begin{tabular}{@{}llrr@{}}
\toprule
Dataset & Method & Selected setting & Test Active \\
\midrule
ETTh1 & TSAE      & shared $\lambda=6.0$, private $\lambda=0.30$ & 231.4$\pm$2.7 \\
      & L1        & $\lambda=2.0$                               & 252.8$\pm$10.3 \\
      & Gated     & $\lambda=6.0$                               & 244.4$\pm$7.4 \\
      & TopK      & $k=320$                                     & 264.9$\pm$5.7 \\
      & JumpReLU  & $\lambda{=}10^{-4}$, threshold${=}0.10$     & 269.8$\pm$11.4 \\
\midrule
ETTh2 & TSAE      & shared $\lambda=3.0$, private $\lambda=0.15$ & 250.9$\pm$13.6 \\
      & L1        & $\lambda=1.5$                               & 195.2$\pm$11.4 \\
      & Gated     & $\lambda=5.0$                               & 212.7$\pm$6.0 \\
      & TopK      & $k=256$                                     & 218.3$\pm$1.0 \\
      & JumpReLU  & $\lambda{=}10^{-4}$, threshold${=}0.08$     & 311.0$\pm$39.9 \\
\midrule
ETTm1 & TSAE      & shared $\lambda=5.0$, private $\lambda=0.25$ & 204.9$\pm$11.5 \\
      & L1        & $\lambda=1.5$                               & 258.0$\pm$11.9 \\
      & Gated     & $\lambda=5.0$                               & 246.9$\pm$5.8 \\
      & TopK      & $k=256$                                     & 223.7$\pm$2.4 \\
      & JumpReLU  & $\lambda{=}10^{-4}$, threshold${=}0.09$     & 269.9$\pm$12.8 \\
\midrule
Weather & TSAE    & shared $\lambda=1.8$, private $\lambda=0.09$ & 216.7$\pm$3.2 \\
        & L1      & $\lambda=1.0$                               & 187.8$\pm$16.3 \\
        & Gated   & $\lambda=2.0$                               & 285.1$\pm$8.8 \\
        & TopK    & $k=320$                                     & 261.1$\pm$14.4 \\
        & JumpReLU& $\lambda{=}10^{-4}$, threshold${=}0.08$     & 236.0$\pm$0.7 \\
\bottomrule
\end{tabular}
\caption{Validation-selected operating points with test-split active counts.
TSAE runs use $\lambda_{\mathrm{temp}}=0.5$, $\lambda_{\mathrm{fc}}=0.1$, and
$\lambda_{\mathrm{balance}}=0.003$.}
\label{tab:appendix-v10-hparams}
\end{table}

All ETT test means in Table~\ref{tab:appendix-v10-hparams} fall within the
prespecified activity window. On Weather, the L1 test mean drifts slightly
below the lower boundary (187.8), while the remaining methods remain
in-window. This protocol controls realized activity per token; it does not
equalize the total parameter count of TSAE's routed dictionaries and flat
baselines.

\subsubsection*{Calibration selection curves}

\begin{figure}[!htbp]
\centering
\includegraphics[width=0.95\textwidth]{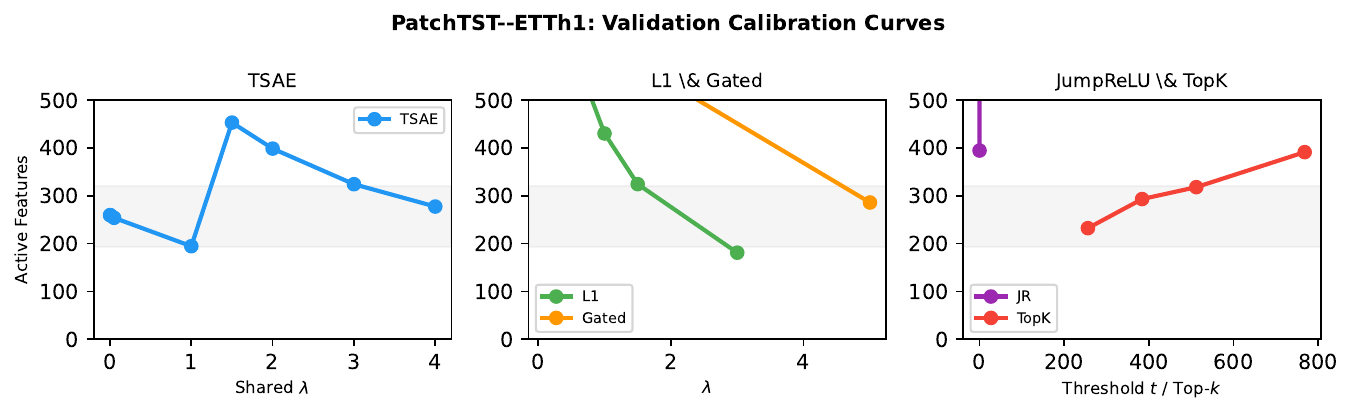}
\caption{Validation calibration curves on PatchTST--ETTh1. Horizontal axis:
sparsity parameter; vertical axis: measured active features per token. Gray
band: admissible [192,320] window.}
\label{fig:appendix-calib-selection}
\end{figure}

Figure~\ref{fig:appendix-calib-selection} shows how the measured active
count responds to the sparsity parameter for each method. The gray band marks
the [192,320] admissible window. For TSAE, the active count decreases
monotonically as the shared sparsity coefficient increases; the selected
operating points are those closest to 256. For L1 and Gated the relationship
is similarly monotonic. JumpReLU and TopK have narrower tuning ranges.

\subsubsection*{Temporal coefficient sensitivity}

We swept $\lambda_{\mathrm{temp}}$ on PatchTST--ETTh1 (single seed,
$\lambda_{\mathrm{shared}}=5.0$, $\lambda_{\mathrm{priv}}=0.2$).
Table~\ref{tab:appendix-temporal} reports the results. Below 0.01, the
transition metric is effectively unchanged. At 0.5 it decreases by about 7\%
relative to zero while the realized activity remains near the target. The
single-seed curve continues to change at 1.0 together with activity, so it is a
sensitivity diagnostic rather than an independent optimality claim. The formal
comparison uses the validation-selected coefficient 0.5.

\begin{table}[!htb]
\centering
\footnotesize
\setlength{\tabcolsep}{2.8pt}
\begin{tabular}{@{}lrrrr@{}}
\toprule
$\lambda_{\mathrm{temp}}$ & Active & Recon.$\downarrow$ & NFRE$\downarrow$ & Trans.$\downarrow$ \\
\midrule
0       & 280 & 12.3 & 5.1 & 0.0908 \\
0.0001  & 280 & 12.3 & 5.1 & 0.0909 \\
0.001   & 280 & 12.3 & 5.1 & 0.0908 \\
0.01    & 278 & 12.3 & 5.1 & 0.0901 \\
0.05    & 276 & 12.2 & 5.1 & 0.0903 \\
0.1     & 273 & 12.2 & 5.1 & 0.0895 \\
\textbf{0.5} & \textbf{255} & \textbf{11.7} & \textbf{4.9} & \textbf{0.0844} \\
1.0     & 236 & 11.0 & 4.5 & 0.0787 \\
0.5 (3-seed) & 268$\pm$11 & 11.7$\pm$0.1 & 4.9$\pm$0.1 & 0.086$\pm$0.003 \\
\bottomrule
\end{tabular}
\caption{Temporal coefficient sweep (single seed unless noted). The bold row
is the validation-selected formal coefficient. Recon.\ and NFRE are
$\times10^{3}$.}
\label{tab:appendix-temporal}
\end{table}

\subsubsection*{Private sparsity sensitivity}

We swept TSAE's private $\lambda$ on PatchTST--ETTh1 (single seed,
$\lambda_{\mathrm{shared}}=5.0$). A larger coefficient ($\lambda_{\mathrm{priv}}
\approx 1.0$) produced 13.8\% dead features. Lowering the coefficient to 0.05--0.2 reduces
the dead ratio in this fixed-setting diagnostic while improving reconstruction
and NFRE. The sweep is illustrative: final formal settings are selected
separately for each dataset under the active-matched protocol.

\begin{table}[!htb]
\centering
\footnotesize
\setlength{\tabcolsep}{2.8pt}
\begin{tabular}{@{}lrrrr@{}}
\toprule
$\lambda_{\mathrm{priv}}$ & Active & Recon.$\downarrow$ & NFRE$\downarrow$ & Dead\% \\
\midrule
0.02 & 500 & 10.7 & 4.4 & 6.2 \\
0.05 & 489 & \textbf{9.4} & \textbf{3.8} & 2.3 \\
0.1  & 456 & 9.5 & 3.9 & 0.6 \\
\textbf{0.2} & \textbf{416} & 10.2 & 4.2 & 1.0 \\
0.5  & 392 & 11.0 & 4.4 & 7.0 \\
1.0  & 386 & 11.3 & 4.5 & 13.8 \\
\bottomrule
\end{tabular}
\caption{Private sparsity sensitivity at a fixed shared coefficient (single
seed). Bold marks the illustrative low-dead-feature range; formal comparisons
use dataset-specific validation-selected values.}
\label{tab:appendix-private}
\end{table}

\subsection{Per-Dataset Active-Matched Achievement}
\label{sec:appendix-active-achievement}

The main-text table compresses the formal comparison. The four tables below
expand it per dataset, showing whether each method's mean test-split active
count falls within the prespecified [192,320] window. The ``In window''
column provides a direct check of the active-matched protocol.

\subsubsection*{ETTh1}

\begin{table}[!htb]
\centering
\footnotesize
\setlength{\tabcolsep}{2.2pt}
\begin{tabular}{@{}lrrrrrr@{}}
\toprule
Method & Active & In 192--320? & Recon.$\downarrow$ & NFRE$\downarrow$ & Dead\% & Trans.$\downarrow$ \\
\midrule
TSAE      & 231$\pm$3   & \checkmark & \textbf{13.4} & \textbf{5.10} & 8.0  & \textbf{0.073} \\
L1        & 253$\pm$10  & \checkmark & 13.6 & 5.33 & \textbf{0.0}  & 0.181 \\
Gated     & 244$\pm$7   & \checkmark & 16.4 & 6.39 & \textbf{0.0}  & 0.144 \\
TopK      & 265$\pm$6   & \checkmark & 30.5 & 10.84& 0.1  & 0.143 \\
JumpReLU  & 270$\pm$11  & \checkmark & 20.7 & 8.09 & \textbf{0.0}  & 0.156 \\
\bottomrule
\end{tabular}
\caption{ETTh1: all five methods within the [192,320] window. TSAE has the
best Recon, NFRE, and Transition rate.}
\label{tab:appendix-active-etth1}
\end{table}

\subsubsection*{ETTh2}

\begin{table}[!htb]
\centering
\footnotesize
\setlength{\tabcolsep}{2.2pt}
\begin{tabular}{@{}lrrrrrr@{}}
\toprule
Method & Active & In 192--320? & Recon.$\downarrow$ & NFRE$\downarrow$ & Dead\% & Trans.$\downarrow$ \\
\midrule
TSAE      & 251$\pm$14  & \checkmark & \textbf{6.04} & \textbf{2.70} & 4.3  & \textbf{0.050} \\
L1        & 195$\pm$11  & \checkmark & 7.99 & 3.66 & 2.5  & 0.106 \\
Gated     & 213$\pm$6   & \checkmark & 8.19 & 3.76 & 10.3 & 0.055 \\
TopK      & 218$\pm$1   & \checkmark & 9.88 & 4.56 & 14.3 & 0.063 \\
JumpReLU  & 311$\pm$40  & \checkmark & 10.9 & 4.79 & \textbf{4.1}  & 0.064 \\
\bottomrule
\end{tabular}
\caption{ETTh2: all five methods are within the [192,320] window. TSAE is the
best on Recon, NFRE, and Transition.}
\label{tab:appendix-active-etth2}
\end{table}

\subsubsection*{ETTm1}

\begin{table}[!htb]
\centering
\footnotesize
\setlength{\tabcolsep}{2.2pt}
\begin{tabular}{@{}lrrrrrr@{}}
\toprule
Method & Active & In 192--320? & Recon.$\downarrow$ & NFRE$\downarrow$ & Dead\% & Trans.$\downarrow$ \\
\midrule
TSAE      & 205$\pm$12  & \checkmark & \textbf{10.4} & \textbf{5.00} & 3.1  & \textbf{0.047} \\
L1        & 258$\pm$12  & \checkmark & 11.4 & 5.50 & \textbf{0.0}  & 0.135 \\
Gated     & 247$\pm$6   & \checkmark & 13.5 & 6.58 & 0.0  & 0.089 \\
TopK      & 224$\pm$2   & \checkmark & 19.9 & 8.02 & 0.6  & 0.083 \\
JumpReLU  & 270$\pm$13  & \checkmark & 17.1 & 8.63 & \textbf{0.0}  & 0.092 \\
\bottomrule
\end{tabular}
\caption{ETTm1: all five methods within the [192,320] window. TSAE has the
best Recon, NFRE, and Transition rate.}
\label{tab:appendix-active-ettm1}
\end{table}

\subsubsection*{Weather}

\begin{table}[!htb]
\centering
\footnotesize
\setlength{\tabcolsep}{2.2pt}
\begin{tabular}{@{}lrrrrrr@{}}
\toprule
Method & Active & In 192--320? & Recon.$\downarrow$ & NFRE$\downarrow$ & Dead\% & Trans.$\downarrow$ \\
\midrule
TSAE      & 217$\pm$3   & \checkmark & \textbf{1.70} & \textbf{2.52} & 19.5 & \textbf{0.019} \\
L1        & 188$\pm$16  & $\times$ & \underline{1.85} & 3.37 & \textbf{0.0}  & 0.055 \\
Gated     & 285$\pm$9   & \checkmark & 2.10 & \underline{2.84} & \underline{0.0} & 0.046 \\
TopK      & 261$\pm$14  & \checkmark & 3.84 & 4.64 & 7.3 & 0.038 \\
JumpReLU  & 236$\pm$1   & \checkmark & 4.49 & 6.21 & 0.1 & \underline{0.031} \\
\bottomrule
\end{tabular}
\caption{Weather: the validation-selected settings yield in-window test means
for TSAE, Gated, TopK, and JumpReLU. L1 is slightly below the boundary at 188,
despite its validation point lying in-window. TSAE has the best Recon, NFRE,
and Transition.}
\label{tab:appendix-active-weather}
\end{table}

This breakdown shows that validation-based active matching is generally
well-controlled across the datasets. On Weather, L1's selected validation point
is in-window, while its three-seed test mean is 188; the other four methods
remain in-window. TSAE retains the best fidelity and temporal-stability metrics
at a comparable realized per-token activity level.

\subsection{Cross-Backbone Extensions}
\label{sec:appendix-cross-backbone}

\subsubsection*{iTransformer--ETTh1 formal comparison}

We ran the same active-matched protocol on frozen iTransformer representations
from ETTh1, using a 1024-feature per-token budget and the same 20-epoch,
30,000/15,000-token budget. Table~\ref{tab:appendix-itransformer} reports
three-seed test-split means.

\begin{table}[!htb]
\centering
\footnotesize
\setlength{\tabcolsep}{2.8pt}
\begin{tabular}{@{}lrrrr@{}}
\toprule
Method & Active & Recon.$\downarrow$ & NFRE$\downarrow$ \\
\midrule
TSAE      & 253$\pm$3  & \textbf{14.5} & \textbf{3.01} \\
L1        & 282$\pm$1  & 18.1 & 4.14 \\
Gated     & 316$\pm$1  & 17.3 & 3.20 \\
TopK      & 212$\pm$0  & 30.1 & 5.24 \\
JumpReLU  & 269$\pm$3  & 18.8 & 3.44 \\
\bottomrule
\end{tabular}
\caption{iTransformer--ETTh1 active-matched results (three-seed means).
Recon.\ and NFRE are $\times10^{3}$.}
\label{tab:appendix-itransformer}
\end{table}

TSAE yields the lowest reconstruction and NFRE on the iTransformer backbone,
consistent with the PatchTST headline result.

\subsubsection*{DLinear--ETTh1 formal comparison}

Table~\ref{tab:appendix-dlinear} reports the same protocol on frozen
DLinear representations from ETTh1.

\begin{table}[!htb]
\centering
\footnotesize
\setlength{\tabcolsep}{2.8pt}
\begin{tabular}{@{}lrrrr@{}}
\toprule
Method & Active & Recon.$\downarrow$ & NFRE$\downarrow$ \\
\midrule
TSAE      & 215$\pm$10 & 9.09 & 7.91 \\
L1        & 250$\pm$1  & 9.17 & 8.74 \\
Gated     & 314$\pm$1  & \textbf{7.71} & \textbf{6.88} \\
TopK      & 256$\pm$0  & 9.90 & 8.27 \\
JumpReLU  & 266$\pm$4  & 7.99 & 7.02 \\
\bottomrule
\end{tabular}
\caption{DLinear--ETTh1 active-matched results (three-seed means).
Recon.\ and NFRE are $\times10^{3}$.}
\label{tab:appendix-dlinear}
\end{table}

On DLinear, Gated SAE has the lowest Reconstruction and NFRE; TSAE is
third on NFRE and second on Reconstruction. The fidelity advantage does not
automatically transfer to every architecture---backbone representation
geometry matters.

\subsubsection*{Multi-backbone component ablation}

Tables~\ref{tab:abl-patchtst}--\ref{tab:abl-transformer} report the
component ablation across four backbones on three datasets. Each table shows
three-seed means with bold marking the best value per column per dataset.
The ablation uses a fixed $\lambda_{\mathrm{shared}}=5.0$,
$\lambda_{\mathrm{priv}}=0.2$, $\lambda_{\mathrm{temp}}=0.5$ configuration
for TSAE. The No temporal variant is included only for PatchTST; all three
other backbones use representation types without ordered temporal adjacency.

\begin{table}[!htb]
\centering
\footnotesize
\setlength{\tabcolsep}{2.4pt}
\begin{tabular}{@{}llrrrrr@{}}
\toprule
Data & Variant & Active & Recon.$\downarrow$ & NFRE$\downarrow$ & Dead\% & Trans.$\downarrow$ \\
\midrule
ETTh1 & Full      & 268 & \textbf{11.7} & \textbf{4.9} & 6.3  & \textbf{0.086} \\
      & No routing & 252 & 13.4 & 5.5 & \textbf{0.0}  & 0.145 \\
      & No gating  & 108 & 18.1 & 7.9 & \textbf{0.0}  & 0.083 \\
      & No temporal& 279 & 12.3 & 5.1 & 6.3  & 0.091 \\
ETTh2 & Full      & 203 & \textbf{2.2}  & \textbf{3.4} & 10.9 & \textbf{0.036} \\
      & No routing & 178 & 2.4  & 3.6 & 13.1 & 0.050 \\
      & No gating  & 46  & 3.7  & 5.7 & 23.3 & 0.027 \\
      & No temporal& 234 & 2.8  & 4.2 & 13.3 & 0.038 \\
ETTm1 & Full      & 218 & \textbf{9.1}  & \textbf{5.2} & 3.8  & \textbf{0.049} \\
      & No routing & 192 & 9.8  & 5.7 & 0.1  & 0.071 \\
      & No gating  & 79  & 14.1 & 8.5 & \textbf{0.0}  & 0.047 \\
      & No temporal& 261 & 10.3 & 5.8 & 6.2  & 0.052 \\
\bottomrule
\end{tabular}
\caption{PatchTST component ablation (three-seed means). Bold marks the best
per column per dataset.}
\label{tab:abl-patchtst}
\end{table}

\begin{table}[!htb]
\centering
\footnotesize
\setlength{\tabcolsep}{2.4pt}
\begin{tabular}{@{}llrrrr@{}}
\toprule
Data & Variant & Active & Recon.$\downarrow$ & NFRE$\downarrow$ & Dead\% \\
\midrule
ETTh1 & Full      & 399 & \textbf{13.7} & \textbf{3.3} & 0.0 \\
      & No routing & 309 & 23.3 & 4.3 & 0.0 \\
      & No gating  & 188 & 34.7 & 7.8 & 0.0 \\
ETTh2 & Full      & 360 & \textbf{22.9} & \textbf{1.1} & 0.0 \\
      & No routing & 284 & 31.6 & 1.4 & 0.0 \\
      & No gating  & 187 & 45.7 & 1.9 & 0.0 \\
ETTm1 & Full      & 383 & \textbf{14.2} & \textbf{4.2} & 0.0 \\
      & No routing & 290 & 23.1 & 5.8 & 0.0 \\
      & No gating  & 154 & 34.4 & 10.1 & 0.0 \\
\bottomrule
\end{tabular}
\caption{iTransformer component ablation. Transition rate not applicable.}
\label{tab:abl-itransformer}
\end{table}

\begin{table}[!htb]
\centering
\footnotesize
\setlength{\tabcolsep}{2.4pt}
\begin{tabular}{@{}llrrrr@{}}
\toprule
Data & Variant & Active & Recon.$\downarrow$ & NFRE$\downarrow$ & Dead\% \\
\midrule
ETTh1 & Full      & 219 & \textbf{21.3} & \textbf{9.8}  & 2.4 \\
      & No routing & 165 & 25.7 & 11.6 & \textbf{0.0} \\
      & No gating  & 121 & 43.5 & 21.9 & \textbf{0.0} \\
ETTh2 & Full      & 153 & \textbf{7.9}  & \textbf{7.3}  & 5.9 \\
      & No routing & 117 & 8.0  & 7.2  & \textbf{0.0} \\
      & No gating  & 87  & 15.0 & 13.2 & \textbf{0.0} \\
ETTm1 & Full      & 139 & \textbf{19.5} & \textbf{13.1} & 4.9 \\
      & No routing & 104 & 21.6 & 15.0 & \textbf{0.0} \\
      & No gating  & 63  & 35.5 & 24.3 & \textbf{0.0} \\
\bottomrule
\end{tabular}
\caption{DLinear component ablation.}
\label{tab:abl-dlinear}
\end{table}

\begin{table}[!htb]
\centering
\footnotesize
\setlength{\tabcolsep}{2.4pt}
\begin{tabular}{@{}llrrrr@{}}
\toprule
Data & Variant & Active & Recon.$\downarrow$ & NFRE$\downarrow$ & Dead\% \\
\midrule
ETTh1 & Full      & 322 & \textbf{13.6} & \textbf{3.5} & 3.2 \\
      & No routing & 281 & 18.7 & 5.0 & \textbf{0.0} \\
      & No gating  & 126 & 38.1 & 12.8 & \textbf{0.0} \\
ETTh2 & Full      & 287 & \textbf{4.2}  & \textbf{4.6} & 8.3 \\
      & No routing & 221 & 5.6  & 6.1 & \textbf{8.7} \\
      & No gating  & 74  & 12.7 & 15.8 & \textbf{0.0} \\
ETTm1 & Full      & 289 & \textbf{8.8}  & \textbf{5.0} & 8.6 \\
      & No routing & 209 & 10.9 & 6.4 & \textbf{3.3} \\
      & No gating  & 83  & 28.6 & 18.1 & \textbf{0.0} \\
\bottomrule
\end{tabular}
\caption{Transformer component ablation.}
\label{tab:abl-transformer}
\end{table}

Across this fixed-configuration matrix, removing routing generally degrades
reconstruction and removing gating lowers realized activity substantially. The
direction of these component effects is observed across several representation
geometries, but the variants are not activity-matched; these tables are
therefore supporting component diagnostics rather than a universal
cross-backbone performance comparison.

\subsection{Reproducibility}
\label{sec:appendix-repro}

All experiments use frozen forecasting backbones trained for 10 epochs with
seed 2024. SAE training uses 20 epochs, batch size 1024, learning rate
$10^{-3}$ (Adam), on at most 30,000 training tokens and 15,000 test tokens.
For the PatchTST headline comparison, TSAE requests
$K_{\mathrm{sh}}=1792$ shared and $K_{\mathrm{priv}}=256$ private features per
variable. A resolver enforces $K_{\mathrm{global}}\leq4096$, so the realized
shared/private allocation can depend on the variable count; flat baselines use
a single 2048-wide dictionary. PatchTST TSAE runs additionally use
$\lambda_{\mathrm{fc}}=0.1$ for two refinement epochs with at most 80 batches
per epoch; baselines do not use this TSAE-specific stage. The iTransformer and
DLinear extensions use their separately reported 1024-feature budgets. Formal
results are averaged over three independent SAE seeds $\{2024,2025,2026\}$.
Experiments were run on NVIDIA RTX 4090 (24\,GB) and RTX 4080 SUPER (32\,GB)
with PyTorch 2.10+cu128 and CUDA 12.8. Complete run-level results,
calibration manifests, and figure-generation scripts are included in the
released artifact.